\pdfoutput=1
\documentclass{article}
\PassOptionsToPackage{numbers,round,sort&compress}{natbib}
\usepackage[accepted]{icml_2025}
\setcitestyle{numbers,round,comma}
\usepackage[utf8]{inputenc}
\usepackage[T1]{fontenc}
\usepackage{lmodern}
\usepackage{microtype}
\usepackage{amsmath,amssymb}
\usepackage{graphicx}
\usepackage{enumitem}
\usepackage{booktabs,array}



\usepackage[hidelinks]{hyperref}
\usepackage{xcolor}
\usepackage{orcidlink}
\providecommand{\tightlist}{\setlength{\itemsep}{0pt}\setlength{\parskip}{0pt}}

\hypersetup{
    colorlinks=true,
    urlcolor=blue!60!black,   % Clean, readable dark blue for URLs
    linkcolor=blue!60!black,  % Internal section/figure links
    citecolor=blue!60!black   % Citation links
}

\makeatletter
\renewcommand{\ICML@appearing}{} % Leave empty {} if you want no notice at all
\makeatother

\renewenvironment{abstract}{%
  \begin{center}
    {\large\bfseries Abstract}\par
    \vspace{1.0em} % Adjust this value (e.g., 0.8em, 1.0em, 1.5em) for more or less space
    \begin{minipage}{0.87\textwidth}
      \small
}{%
    \end{minipage}
  \end{center}
  \vspace{0.25in}
}

\icmltitlerunning{Loop-Count Schedules in Looped Transformers: An Empirical Study}

\begin{document}

\twocolumn[
\icmltitle{Loop-Count Schedules in Looped Transformers: An Empirical Study}

\begin{icmlauthorlist}
\icmlauthor{Konstantin Kolokolov}{}\orcidlink{0009-0000-2971-2325}
\end{icmlauthorlist}

\icmlkeywords{Looped Transformers, Test-Time Compute, Machine Learning}

\vskip 0.3in

\begin{abstract}
Looped transformers apply one block of layers several times with shared weights, trading parameters for repeated computation. We study how the schedule of the loop iteration count R during training determines whether each extra pass acts as a full layer or as a small correction, using a controlled ladder of 46M--110M-parameter models trained on 2B tokens with learned residual gates on every loop pass and three schedules: R fixed to 2, R sampled uniformly from \(\{1, \dots, 6\}\), and R sampled from \(\{2, 3, 4\}\). We find that the lower bound of the sampling range is the decisive choice. If the sampled range includes R=1, so that some training steps run the block only once and compute the loss on that single pass, the gates collapse to 0.1--0.5 and each additional pass becomes a small correction to the first; if the range starts at R=2, so that every training step runs the block at least twice, the gates stay at 0.6--0.9 and each pass keeps the strength of a full layer, while the model can still be run at any R from 2 up to twice the largest trained value with validation loss within a few hundredths of a nat of its best. The \(\{2, 3, 4\}\) schedule gives the best looped model, 2.909 nats against 2.951 for the parameter-matched plain model, and degrades by only 0.038 nats when run at twice its trained depth. We further show that under a fixed R the direction of the residual stream converges geometrically beyond the trained depth while the loss diverges, which refines a recent finding that static-depth training does not converge, and that reusing the first pass's keys and values on later passes costs 0.01--0.03 nats in every regime while making the per-token iteration uniformly contractive and halving the inference cache. The loop recovers about a third of what real depth provides at equal parameters. The code, trained weights, and per-token traces are released.
\end{abstract}

]

% first-page footnote block (bottom left), in place of \printAffiliationsAndNotice
{\let\thefootnote\relax\footnotetext{\hspace*{-\footnotesep}ORCID: \href{https://orcid.org/0009-0000-2971-2325}{0009-0000-2971-2325}}}

\section{Introduction}\label{sec:intro}

Reusing one block of a transformer across several passes has been proposed by several groups over the past years, first for parameter efficiency and adaptive computation and more recently as a route to latent reasoning. Dehghani et al.~\citep{dehghani2019universal} introduced the Universal Transformer, which applies a single block recurrently with a step embedding and an adaptive halting mechanism; Lan et al.~\citep{lan2020albert} shared weights across all layers of ALBERT to cut parameters at a small cost in quality. A second line of work treated the looped transformer as a machine for iterative algorithms: Yang et al.~\citep{yang2024looped} showed that it learns in-context learning algorithms better than an unrolled stack, and Fan et al.~\citep{fan2024looped} that it generalizes to longer inputs when the number of loops scales with the input.

The current motivation is latent reasoning at language-model scale. Bae et al.~\citep{bae2024relaxed} revisited parameter sharing with per-loop low-rank adapters and, in Mixture-of-Recursions \citep{bae2025mor}, added token-level routing of the recursion depth together with several KV-cache sharing policies. Geiping et al.~\citep{geiping2025scaling} trained a 3.5B-parameter recurrent-depth model with the loop count sampled during training and the input injected at every step, and reported that extra passes at inference improve reasoning benchmarks. Zhu et al.~\citep{zhu2025ouro} pre-trained the Ouro family on 7.7T tokens with an entropy-regularized objective that learns how much depth to allocate per input, attributing the gains to better manipulation of knowledge rather than larger capacity; Marchenko et al.~\citep{marchenko2026closing} reduced that budget to 310B tokens with a from-scratch recipe and a minimal procedure for converting pretrained dense models. Jeddi et al.~\citep{jeddi2026loopformer} trained on trajectories of varying length with a shortcut-consistency objective so that short runs give usable states and long runs refine them, and Wang et al.~\citep{wang2026loopmoe} found that untying the attention parameters across passes of a looped mixture-of-experts helps while the experts stay shared. Huang et al.~\citep{huang2026optimizers} reframed the loop transition as an optimizer step with a learned, monotonically decreasing step size, and observed that perplexity understates the generative gains of looped models; Zeitoun et al.~\citep{zeitoun2026hyperloop} widened the residual stream into several streams that are mixed after each loop and matched depth-matched baselines with about half the parameters. Alongside these are studies of where recurrent compute is best allocated \citep{lin2026allocating,zhuang2026recurrence} and of how to decode looped models efficiently \citep{liu2026loopahead,liu2026decoding}. Ladhak et al.~\citep{ladhak2026elr} and Fishman et al.~\citep{fishman2026dory} carried the recipe to 4B parameters and above, with loop counts mixed within each optimizer update over \(\{1, \dots, 4\}\), a gated state update with a learned step size, and only the last loop's KV cache kept at inference; all three choices have small-scale counterparts in this paper, and the lower bound of 1 in their sampled range is the setting our results identify as the damped regime.

Two questions about the loop's dynamics have received less attention than its benchmarks. The first is whether the iteration converges. Dong et al.~\citep{dong2021attention} showed that a stack of pure self-attention layers loses rank doubly exponentially and that the feed-forward sublayer is what prevents the collapse, so the two halves of a block pull in opposite directions when iterated. Logan \citep{logan2026fixedpoint} measured per-token fixed-point convergence in depth-recurrent transformers with the KL divergence between consecutive passes' output distributions and found models saturating at their training-mean depth; Pappone et al.~\citep{pappone2025twoscale} observed that the loop steps of such a model become smaller and increasingly orthogonal over training and built an early-exit rule on the second-order difference of the step size. Most directly, Prairie et al.~\citep{prairie2026benice}, in independent work developed in parallel with this study, trained looped models at 140M--1.3B parameters with and without a stochastically sampled depth and showed, using the relative step of the residual stream, that stochastic-depth training drives the representations toward a common limit across passes whereas static-depth training does not; they use this property to share the KV cache of the final pass backward into the earlier ones after training, zero-shot for models pretrained with sampled depth and with a short finetuning stage otherwise. The second question is what reusing keys and values across passes does. Jaegle et al.~\citep{jaegle2021perceiver} iterated a latent array against a fixed input array in the Perceiver, which is the earliest form of a loop with a static memory; Bae et al.~\citep{bae2025mor} measured the quality cost of reusing the first recursion's cache; Brandon et al.~\citep{brandon2024cla} and Sun et al.~\citep{sun2024yoco} share caches across distinct layers rather than across passes. In our own earlier work on contextual biasing for speech recognition \citep{kolokolov2024selfconsistent} we iterated a transducer joiner to self-consistency against a fixed set of hint embeddings, a loop whose only feedback path is the attention query, and observed convergence within three iterations; the contrast between that fast, bounded iteration and the slow contraction of a looped language-model block is the origin of the static-KV variant studied here.

What this literature leaves open is which property of the loop-count schedule produces the convergent regime, what the model gives up to obtain it, and whether the two can be separated. The present paper answers these questions in a setting small enough to change one thing at a time, and reads the answer off two instruments that the larger studies do not report: the learned residual gates on each loop pass, and per-token traces of the residual stream's direction and norm across passes.

\section{Model and schedule settings}\label{sec:model}

\begin{figure*}[t]
\centering
\includegraphics[width=\textwidth]{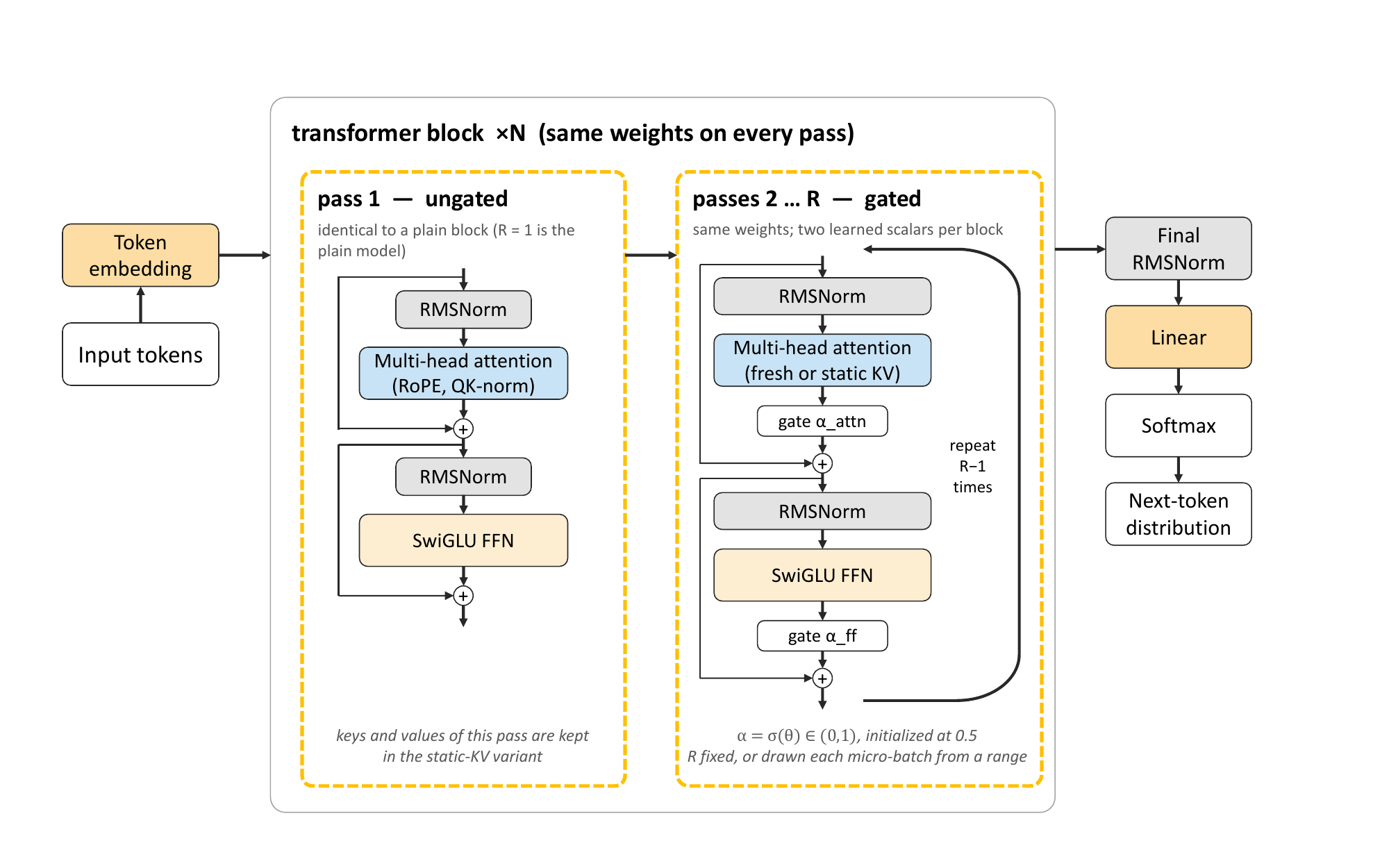}
\caption{The model. Tokens are embedded and pass through N stacked blocks; each block runs once ungated and then $R-1$ more times with the same weights and gated residual updates, after which a final normalization, the tied output projection and a softmax give the next-token distribution. R=1 is the plain model. In passes 2 to R the gates $\alpha_\mathrm{attn}$ and $\alpha_\mathrm{ff}$, shown as small boxes after each sublayer, scale that sublayer's update to the residual stream.}
\label{fig:model}
\end{figure*}

We use a decoder-only transformer whose layers are standard pre-norm blocks: RMSNorm \citep{zhang2019rmsnorm} before each sublayer, multi-head attention with rotary position embeddings \citep{su2021roformer} and query-key normalization \citep{henry2020qknorm}, and a SwiGLU feed-forward network \citep{shazeer2020glu}; token embeddings are tied to the output projection, and the output head is a final RMSNorm, a linear layer and a softmax over the vocabulary. The plain model applies each block once. The looped model applies each block R times with shared weights, as in Figure~\ref{fig:model}.

The looping version of the transformer block is organized the following way. The first pass is ungated and identical to a plain block. All following loop passes (from 2 to R) add gated residual updates with separate learned scalars on the attention and FFN paths:
\[
\begin{aligned}
x &\leftarrow x + \alpha_{\text{attn}}\,\mathrm{Attn}(\mathrm{Norm}_1(x)), \\
x &\leftarrow x + \alpha_{\text{ff}}\,\mathrm{FFN}(\mathrm{Norm}_2(x)).
\end{aligned}
\]
Here $x$ is the residual stream, the per-token hidden vector of width $d$ that the skip connections carry through the block; every sublayer reads it through its pre-norm and adds its output back into it. Norm denotes RMSNorm throughout; there is no LayerNorm anywhere in the model.

Each gate is parameterized as $\alpha = \sigma(\theta)$, with $\sigma$ the logistic sigmoid and $\theta$ a trainable parameter, so it lies in (0, 1); both are initialized at 0.5 ($\theta = 0$), excluded from weight decay, and logged every step. A gate near 1 means the pass writes at the full strength of a plain layer; a gate near 0 means the pass is nearly skipped. At R=1 the block is a plain block, which we use as a correctness test. All blocks loop with the same R; there is no step embedding, no input injection, and no truncated backpropagation. The gates are the minimal form of the learned step size that Huang et al.~\citep{huang2026optimizers} make input-dependent and monotone across loops.

\textbf{Loop iteration schedules.} The number of passes R is a training-time choice, and we compare three ways of making it. In all three, R applies to every block of the model on a given forward pass, the loss is the ordinary next-token cross-entropy on the output of the final pass, and gradients flow through all passes.

R counts every pass, so R=1 is the ungated first pass alone, i.e.~a plain block.

\begin{itemize}[noitemsep, topsep=2pt]
\item
  \textbf{Fixed R=2.} Every training forward pass runs each block exactly twice.
\item
  \textbf{R sampled from \(\{1, \dots, 6\}\).} R is drawn uniformly from 1 to 6, so some training passes run each block only once and compute the loss on that single pass; the mean number of passes is 3.5.
\item
  \textbf{R sampled from \(\{2, 3, 4\}\).} R is drawn uniformly from 2 to 4, so every training pass runs each block at least twice; the mean number of passes is 3. Relative to the previous schedule this changes the lower bound of the range (and, to keep the compute per step comparable, the upper bound).
\end{itemize}

In the sampled schedules R is drawn independently for each micro-batch rather than once per optimizer step; with gradient accumulation an optimizer step therefore averages gradients taken at several different depths, so the update the weights receive is already an average over the schedule rather than a single draw (the micro-batch size and accumulation factor are set per configuration to fit memory, with the tokens per optimizer step held fixed). Every model, whatever its training schedule, is evaluated at \(R \in \{1, \dots, 6, 8\}\), which covers the trained ranges and extends beyond them.

\textbf{Forward static KV.} Attention forms queries Q, keys K and values V by three linear projections of the normalized residual stream; each is a tensor of shape (sequence length $\times$ head dimension) per head, and we write KV for the pair. Under static KV, keys and values are projected on the first pass and reused on later passes; only the query is recomputed from the updated state. The attention output on a later pass is a convex combination of fixed value vectors steered by a fresh query, the KV projections are skipped on later passes, the inference cache is that of a plain model with the same block count, and token i's update no longer depends on the current state of token j. We call this forward sharing, since the first pass's cache serves the later passes (Bae et al.~\citep{bae2025mor} share the first recursion's cache in the same direction); the converse, broadcasting the final pass's cache to all earlier passes after training, as Prairie et al.~\citep{prairie2026benice} do, we call backward sharing. The forward variant transfers to the transformer block the structure of the looped context joiner of \citep{kolokolov2024selfconsistent}, in which a small attention module is iterated against a fixed memory of hint embeddings with only its query changing from one iteration to the next.

\textbf{Convergence metrics.} Fix one block and one token position, and let $s_0$ be that token's residual-stream vector (of dimension d) as it enters the block. Let $s_i$ be the same token's residual vector after the block's $i$-th pass, $i = 1, \ldots, R$, so that $s_1$ is the output of the ungated first pass and $s_R$ is what the next block receives. We record two per-token, per-block quantities for every pass:
\[
\begin{aligned}
&\mathrm{rel}_i = \frac{\lVert s_i - s_{i-1} \rVert_2}{\lVert s_{i-1} \rVert_2}, \\
&\delta_i = 1 - \cos\varphi_i = 1 - \frac{s_i^{\top} s_{i-1}}{\lVert s_i \rVert_2 \, \lVert s_{i-1} \rVert_2}, \\
&i = 1, \dots, R,
\end{aligned}
\]
in fp32, on a fixed validation batch per domain at every checkpoint (16 sequences of 1,024 tokens, or 8 for the 6-block sampled-R runs, whose micro-batch is halved to fit memory). $\mathrm{rel}_i$ is the relative size of the step the pass makes, the metric used by Prairie et al.~\citep{prairie2026benice}; it is sensitive to changes in both the direction and the length of the residual vector. $\delta_i$ measures only the change in direction: $\varphi_i$ is the angle between $s_i$ and $s_{i-1}$, so $\delta_i$ is the cosine distance between consecutive states, 0 when the pass leaves the direction unchanged and 1 when the two are orthogonal. For a pre-norm block only the direction of the residual reaches the next sublayer and the output projection, since each reads Norm(s), so $\delta_i$ is our primary metric and $\mathrm{rel}_i$ is reported for comparison with that work. We summarize both as the median over the tokens of the batch within each block, since the per-token distribution is heavy-tailed (a small set of tokens keeps moving far more than the rest) and a mean would be dominated by them, and report ratios of consecutive medians as the contraction rate. The output-space counterpart is the KL divergence between the next-token distributions read out from consecutive passes,
\vspace{5pt}
\[
\begin{aligned}
&\mathrm{KL}_i = \mathrm{KL}\big(p_i \,\|\, p_{i-1}\big), \\
&p_i = \mathrm{softmax}\big(W\,\mathrm{Norm}(s_i)\big),
\end{aligned}
\]
where Norm is the final RMSNorm and W the tied output projection, so that $p_R$ of the last block is the model's actual next-token distribution and every other $p_i$ is a logit-lens readout of an intermediate state. $\mathrm{KL}_i$ is recorded by the same trace for the runs trained after it was added (the seed-43 runs and the seed-42 \(R \sim \{2, 3, 4\}\) runs); the other seed-42 runs predate its logging.

\section{Experimental setup}\label{sec:setup}

\textbf{Data.} The training corpus is a 2.0B-token mixture of three public sources: FineWeb-Edu (the sample-10BT subset) for general text, the Python portion of codeparrot-clean for code, and FineMath-4+ for mathematical text. Documents longer than 16k tokens are truncated. A 32k-entry BPE tokenizer is trained on 300k documents drawn from the mixture in its training proportions and used for every run. From each source, 1M tokens of documents are held out before training as that domain's validation set, so no validation text is seen in training. The remaining documents are tokenized, separated by a beginning-of-sequence token, and packed into fixed 1024-token training sequences; the training stream draws these sequences from the three domains in the proportions 80\% text, 10\% code, 10\% math, for a total of 2.0B tokens, which is one pass over the prepared data. Validation uses the same packing, and the validation loss is computed separately on each domain's held-out set; the single validation loss we report is the mean of the three per-domain losses weighted by the training proportions, so it corresponds to the loss on a stream drawn like the training data. Per-domain losses appear in Table~\ref{tab:A4} and differ from each other, but nowhere in this paper do the domains differ in their response to the loop beyond single-seed noise, so the text reports the weighted validation loss throughout.

\textbf{Models.} All models share one width and differ only in depth and in whether they loop. The residual stream has a hidden dimension of 768; attention uses 12 heads of 64 dimensions each; the feed-forward sublayer expands to an inner dimension of 2048 in the SwiGLU form, whose two input projections give it the same parameter count as a conventional MLP with an inner dimension of $8/3 \times 768$, so plain and looped models are matched on purpose. The vocabulary has 32,768 entries, and the embedding matrix is tied: the same $32{,}768 \times 768$ matrix that maps a token id to its input vector is reused, transposed, as the output projection that maps the final hidden state to logits over the vocabulary, so the model carries one such matrix rather than two. Each block holds about 7.1M parameters and the tied embedding about 25.2M, so the models have 21.2M, 42.5M and 85.0M parameters outside the embedding (3, 6 and 12 blocks) and 46.4M, 67.6M and 110.1M in total. Looping adds only the gates: two scalars per block.

\textbf{Training.} Every run uses the same optimizer, schedule and token budget, so the only differences between runs are the ones named in \S\ref{sec:model}. We train with AdamW ($\beta_1 = 0.9$, $\beta_2 = 0.95$) at a peak learning rate of $6 \times 10^{-4}$, warmed up linearly over the first 2,000 steps and then decayed with a cosine schedule to 10\% of the peak by the end of training. Weight decay of 0.1 is applied to the weight matrices only, not to normalization parameters, biases or the loop gates. Gradients are clipped to a global norm of 1.0. Computation is in bfloat16 with float32 master weights. Each optimizer step sees 32,768 tokens (32 sequences of 1,024 tokens, accumulated over several micro-batches whose size depends on the configuration's memory footprint), and training runs for 61,000 steps, which is one pass over the 2.0B-token corpus. Checkpoints and validation are taken every 2,000 steps, so the last evaluated checkpoint, from which every number in this paper is taken and which is the one released, is at step 60k. Each run occupies one RTX 5080 (16 GB) for between one and four days depending on depth and loop schedule. Every run uses random seed 42 for initialization and data order. The four 6-block runs that carry the headline comparison, the plain 6-block model and the looped 6-block model under each of the three schedules with fresh KV, were repeated with seed 43; where both seeds exist, Tables 1, 2 and 5 and the text report the two-seed mean, and the two seeds agree to within 0.003 nats at every trained loop count (Appendix A).

\textbf{Runs.} The ladder consists of three plain models with 3, 6 and 12 blocks (plain\_3, plain\_6, plain\_12) and looped versions of the 3- and 6-block models. The 6-block looped model is trained under each of the three loop-count schedules, and each of those both with keys and values recomputed on every pass and with the static-KV variant, giving six looped 6-block runs. The 3-block looped model is trained under the fixed R=2 and the \(R \sim \{1, \dots, 6\}\) schedules, again with both KV policies, giving four looped 3-block runs. That is thirteen runs in total; their names and final numbers are listed in Appendix A.

\textbf{Evaluation.} Every 2,000 training steps we compute the validation loss on each domain's held-out set, first at the loop count used as the headline for that run (R=2 for every looped model, so the headline numbers are comparable), and then again at every \(R \in \{1, \dots, 6, 8\}\), by looping the same weights a different number of times with no retraining; this gives the loss-versus-R curves of \S\ref{sec:lossR} at every checkpoint. At the same checkpoints we run the convergence trace of \S\ref{sec:model} on the fixed validation batch, to 6 passes for the fixed-R models and to 8 for the sampled-R models, and save the per-token arrays. The loop gates are logged at every training step. We also record training throughput in tokens per second and peak GPU memory for each run, since the cost of the variants is part of the comparison.

\section{Problem statement and summary of findings}\label{sec:problem}

Nearly every recent recipe samples the number of passes R during training \citep{geiping2025scaling,zhu2025ouro,marchenko2026closing,jeddi2026loopformer}, and Prairie et al.~\citep{prairie2026benice} show at scale why that matters for inference: stochastic depth makes the residual stream converge across passes, which is what makes the KV cache shareable. We ask which property of the sampling is responsible, what the model gives up to obtain it, and whether the two can be separated.

We answer in a controlled setting. We train plain transformers of 3, 6 and 12 layers and looped versions of the 3- and 6-layer models with learned residual gates on every loop pass, on one 2B-token mixture, and change only the R schedule and the KV policy. The gates turn out to be the observable: their learned values say directly how much work a loop pass does. Three schedules give three answers.
\begin{itemize}
\tightlist
\item
  \textbf{Fixed R=2.} Gates in deep blocks go to $\approx 1$. The loop is an extra layer. The model is 0.032 nats better than the parameter-matched plain model at R=2 and loses between half a nat and a full nat at R=6, depending on the seed.
\item
  \textbf{R sampled from \(\{1, \dots, 6\}\).} Gates collapse to 0.1--0.5. The loop is a damper: each pass is a small correction to a first pass that does most of the work. Loss is flat for R from 2 to 8 and the total gain from extra passes is at most 0.09 nats.
\item
  \textbf{R sampled from \(\{2, 3, 4\}\).} The only change from the previous schedule is that R=1 is never drawn. Gates stay at 0.6--0.9, the loss is the best of all looped models (2.909 at R=3) and flat over the trained range, and the penalty at R=8 is 0.038 nats.
\end{itemize}
The lower bound of the schedule is therefore the switch. Asking the model to be usable after one pass pushes the computation into the first pass and makes every further pass small, which is the convergent regime observed at scale and also the regime in which extra passes have little to add. Excluding R=1 keeps the loop a full layer while sampling over 2--4 keeps it robust there.

Two observations about the loop's dynamics support and refine this. First, the fixed-R loop does converge, in the sense that matters for a pre-norm model: the direction of the residual stream settles geometrically past the trained depth while its norm grows, and the loss diverges because the direction settles toward a state the output projection never saw. The relative residual step is dominated by the norm and reads this as non-convergence. Second, freezing keys and values after the first pass and reusing them forward, the opposite direction to backward cache sharing, costs 0.01--0.03 nats in every regime, makes the per-token iteration uniformly contractive, and halves the inference cache.

The contributions are: (i) the lower bound of the loop-count schedule, not sampling itself, decides whether a looped block learns an extra layer or a damped iteration, with the learned gates as the observable and a one-flag intervention as the test; (ii) a direction-versus-norm account of hidden-state convergence under fixed depth that reconciles geometric settling with diverging loss; (iii) the cost and the dynamics of forward static KV trained from scratch; (iv) a parameter- and FLOP-matched ladder quantifying how much of real depth the loop recovers (35\% at 3 blocks, 24\% at 6). The training code (the gatedloop-transformer package), every configuration in this paper, and all runs and per-token traces are released at \url{https://github.com/kvoyager/gatedloop-transformer}, and the trained weights at \url{https://huggingface.co/konstank/gatedloop-transformer}.

\section{Results}\label{sec:results}

\subsection{The lower bound of the schedule sets the gates}\label{sec:gates}

\begin{figure*}[t]
\centering
\includegraphics[width=0.8\textwidth]{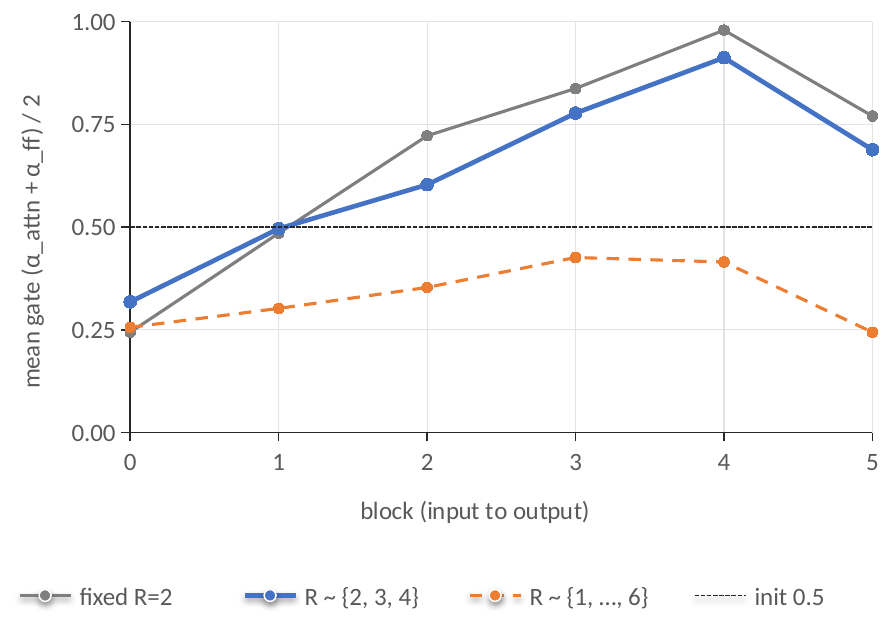}
\caption{The learned loop gates at the last checkpoint (step 60k), one value per block, for the three training schedules. The gate is the scalar $\alpha$ that multiplies a loop pass's update to the residual stream (\S\ref{sec:model}); a value near 1 means the extra pass is used at full strength, a value near 0 means it is nearly skipped. Plotted is the mean of the attention and feed-forward gates of each block, averaged over two seeds, for the 6-block models that recompute keys and values on every pass. Removing R=1 from the sampling range restores the fixed-R profile.}
\label{fig:gates}
\end{figure*}

The gates say directly how much a loop pass does (Figure~\ref{fig:gates}, Table~\ref{tab:1}). Under fixed R=2 the deeper blocks run the second pass at full strength (block 4: $\alpha_\mathrm{attn} = 1.00$, $\alpha_\mathrm{ff} = 0.96$). Under \(R \sim \{1, \dots, 6\}\) every gate drops to 0.1--0.5. Under \(R \sim \{2, 3, 4\}\) the profile returns to within 0.15 of the fixed-R profile with the same shape, ramping from the first block to block 4. The three profiles are the same in both seeds: no block-mean gate differs between the seeds by more than 0.07, against a gap of 0.3--0.5 between the \(R \sim \{1, \dots, 6\}\) profile and the other two. Attention gates exceed FFN gates in the early and middle blocks of every variant, consistent with attention being the contractive sublayer and the FFN the expansive one \citep{dong2021attention}; the last block reverses the order, and the first block loops least everywhere.

\begin{table}[htbp]
\centering
\small
\caption{$\alpha_\mathrm{attn}$ / $\alpha_\mathrm{ff}$ per block at step 60k, 6-block fresh-KV models, mean over seeds 42 and 43. No single gate differs between the two seeds by more than 0.10; the per-seed values are in Appendix A.}
\label{tab:1}
\vspace{3pt}
\begin{tabular}{lccc}
\toprule
block & fixed R=2 & \(R \sim \{1, \dots, 6\}\) & \(R \sim \{2, 3, 4\}\) \\
\midrule
0 & 0.31 / 0.18 & 0.41 / 0.10 & 0.47 / 0.17 \\
1 & 0.54 / 0.43 & 0.31 / 0.30 & 0.47 / 0.52 \\
2 & 0.84 / 0.60 & 0.33 / 0.38 & 0.65 / 0.55 \\
3 & 0.97 / 0.71 & 0.40 / 0.45 & 0.81 / 0.75 \\
4 & 1.00 / 0.96 & 0.46 / 0.37 & 1.00 / 0.83 \\
5 & 0.63 / 0.91 & 0.12 / 0.37 & 0.49 / 0.89 \\
\bottomrule
\end{tabular}
\end{table}

The interpretation is mechanical. A model that must produce a usable prediction after a single pass has to put the computation into that pass, and every later pass can then only be a small correction; a model that is never evaluated at R=1 is free to spread the computation across passes and use each at full strength. The \(\{2, 3, 4\}\) run is the test: the schedule is the only difference from \(R \sim \{1, \dots, 6\}\), and the gates recover. The 3-block models show the same ordering with a milder collapse (mean gate 0.81 fixed, 0.53 under \(R \sim \{1, \dots, 6\}\); Appendix A), plausibly because three blocks leave less room to concentrate work in the first pass.

\subsection{Loss vs inference R}\label{sec:lossR}

\begin{figure*}[t]
\centering
\includegraphics[width=0.9\textwidth]{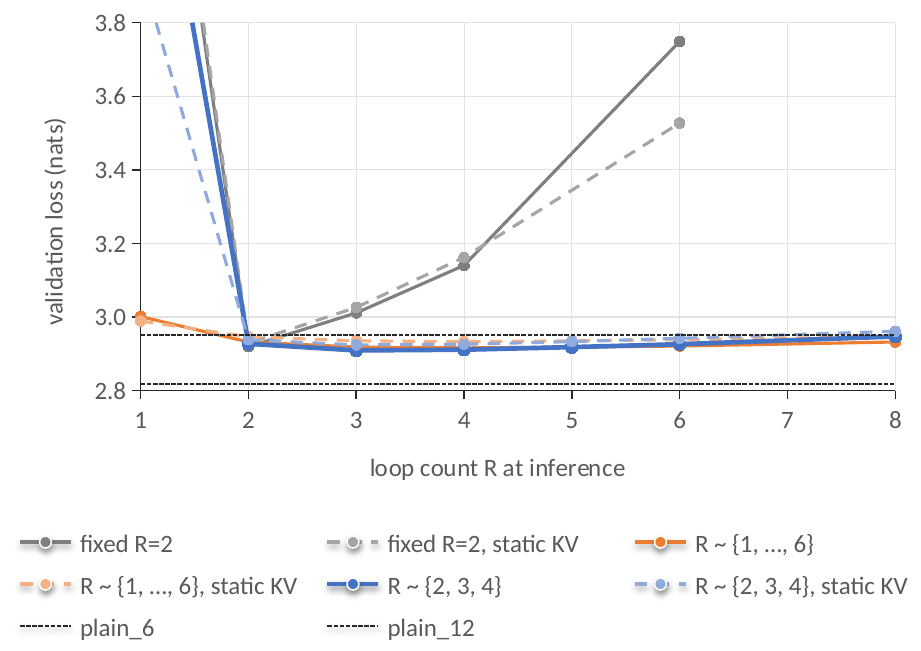}
\caption{Validation loss of the six 6-block looped models as a function of the number of passes R the shared block is applied at inference time. Each curve is one trained model evaluated at every R from 1 to 8 without retraining; the three colours are the three training schedules of \S\ref{sec:model}. Solid lines are models in which attention recomputes its keys and values on every pass from the current state, shown as the mean over two seeds; dashed lines are the static-KV variants, which compute keys and values once on the first pass and reuse them on every later pass, recomputing only the query, each a single run. Horizontal dotted lines are the plain (non-looped) 6-layer model (2.951) and 12-layer model (2.818) for reference. Values above the 3.80 limit of the axis, all at R=1: the fixed-R model is at 4.96 (static KV 5.01) and the \(R \sim \{2, 3, 4\}\) model at 4.60 (static KV 3.94). The best looped value is 2.909, at R=3 for \(R \sim \{2, 3, 4\}\).}
\label{fig:lossR}
\end{figure*}

The three gate profiles correspond to three loss curves (Figure~\ref{fig:lossR}, Table~\ref{tab:2}). The fixed-R model is a two-pass network: 4.96 at R=1, best at R=2, +0.09 at R=3 and +0.83 at R=6, where the two seeds disagree most (4.06 and 3.43), so its loss beyond the trained depth is not a stable property of the schedule. The \(R \sim \{1, \dots, 6\}\) model is flat from R=2 to R=6 (spread 0.016) and usable at R=1 (3.00, against 2.951 for the plain model, so the one-pass regime costs 0.05 nats for having been trained as part of a loop), but the whole gain from R=1 to its best point is 0.085 nats and that best point (2.916 at R=4) only matches the fixed-R model's R=2 (2.919). The \(R \sim \{2, 3, 4\}\) model is the best looped model (2.909 at R=3), flat over its trained range (2.927 / 2.909 / 2.911), degrades by 0.038 at R=8, and is not usable at R=1 (4.60) because it never trained there. Its margin over the fixed-R model is 0.010, small but of the same sign and size in both seeds (0.009 and 0.010) against a seed-to-seed difference of 0.001--0.003 at any trained R. Sampling costs 0.008 (\(R \sim \{2, 3, 4\}\)) and 0.013 (\(R \sim \{1, \dots, 6\}\)) nats at the matched R=2. The optimum is at R=3--4 at every checkpoint from 10k steps on, in both seeds, and the per-domain curves have identical shape.

\begin{table*}[t]
\centering
\small
\caption{Validation loss of the 6-block looped models as a function of R, the number of passes the shared block is applied at inference, for the three training schedules; keys and values are recomputed on every pass. Each entry is the mean over seeds 42 and 43. The two seeds differ by at most 0.003 at any R inside a model's trained range, and by 0.18 (R=1) and 0.63 (R=6) for the fixed-R model outside it; per-seed values are in Appendix A. Bold marks the best R for each model. For reference, the plain 6-layer model scores 2.951 (both seeds) and the plain 12-layer model 2.818.}
\label{tab:2}
\vspace{3pt}
\begin{tabular}{lccccccc}
\toprule
schedule & R=1 & R=2 & R=3 & R=4 & R=5 & R=6 & R=8 \\
\midrule
fixed R=2 & 4.956 & \textbf{2.919} & 3.011 & 3.140 & --- & 3.749 & --- \\
\(R \sim \{1, \dots, 6\}\) & 3.001 & 2.932 & 2.918 & \textbf{2.916} & 2.918 & 2.921 & 2.932 \\
\(R \sim \{2, 3, 4\}\) & 4.602 & 2.927 & \textbf{2.909} & 2.911 & 2.918 & 2.926 & 2.947 \\
\bottomrule
\end{tabular}
\end{table*}

None of the three schedules shows more than 0.09 nats of test-time scaling on language-modeling loss, and all saturate by $R \approx 4$. We read this as a property of the metric and the scale rather than of the loop: Huang et al.~\citep{huang2026optimizers} report looped models whose perplexity trails a dense baseline while their generative accuracy leads, and the reasoning gains of Geiping et al.~\citep{geiping2025scaling}, Zhu et al.~\citep{zhu2025ouro} and Marchenko et al.~\citep{marchenko2026closing} are on tasks where a pass can change a discrete answer. What the schedule buys at no quality cost is the freedom to run the model at any R in its trained range.

\subsection{Direction settles, norm grows}\label{sec:direction}

Prairie et al.~\citep{prairie2026benice} report, using the relative residual step, that stochastic-depth training converges to a fixed point across recurrences and static-depth training does not. Our traces reproduce the relative-step picture and add a distinction. Under fixed R=2 the median direction change $\delta_i$ falls by 0.6--0.8$\times$ per pass in every block out to pass 6 (block 5: 0.047, 0.017, 0.010, 0.006, 0.004 at passes 2--6), while the residual norm grows by about 9\% per pass (Table~\ref{tab:A3}) and the relative step floors at 0.12--0.20; at pass 6 the two account for the floor together, since $\mathrm{rel}^2 \approx (r-1)^2 + 2\delta$ with $r$ the norm ratio gives $0.089^2 + 2 \times 0.0031 = 0.119^2$ against the measured 0.124. In a pre-norm block every sublayer and the output projection read Norm(x), so the direction is the state and the norm is invisible to them: by the metric that matters, the fixed-R loop converges. The loss over the same passes rises from 2.92 to 4.06. The extra passes move the direction a little, consistently, toward a state the output projection never saw; the static-depth failure is in where the iteration converges, not whether.

Under sampled R the two measures agree: the direction settles through pass 8 and the loss is flat through pass 6. The size of the loop step tracks the gates. \(R \sim \{1, \dots, 6\}\) takes steps an order of magnitude smaller than fixed R (block-mean median $\delta_2$: 0.0034 vs 0.020); \(R \sim \{2, 3, 4\}\) takes steps five times larger than that (0.0168) and still contracts geometrically with per-pass ratios of 0.5--0.8 in every block, whereas \(R \sim \{1, \dots, 6\}\) has one block (block 4) that is briefly expansive (ratios 1.09, 1.05 at passes 3--4).

\begin{table*}[t]
\centering
\small
\caption{Upper rows: how much the residual stream changes direction at each successive pass of the loop, measured as $\delta_i$, one minus the cosine of the angle between a token's residual vector after pass $i$ and after pass $i-1$, taken as the median over tokens within each block and then averaged over the six blocks (seed-42 runs). Middle rows: the relative step $\mathrm{rel}_i$ of \S\ref{sec:model}, which includes the change in norm, with the same aggregation and from the same traces. Lower rows: the output-space counterpart, the KL divergence in nats between the next-token distributions read out from the residual after pass $i$ and after pass $i-1$ through the final norm and the tied head, with the same aggregation, for the seed-43 runs of the same three models (most seed-42 traces predate this logging; the seed-43 direction values are within 25\% of those shown). Pass 1 is the ungated first pass, so the table starts at pass 2. 6-block models that recompute keys and values on every pass, evaluated on the FineWeb-Edu validation batch at the last checkpoint (step 60k). Smaller values mean the state, or the prediction, is settling.}
\label{tab:3}
\vspace{3pt}
\begin{tabular}{lccccccc}
\toprule
metric, schedule & p2 & p3 & p4 & p5 & p6 & p7 & p8 \\
\midrule
$\delta$, fixed R=2 & 0.020 & 0.009 & 0.006 & 0.004 & 0.003 & --- & --- \\
$\delta$, \(R \sim \{1, \dots, 6\}\) & 0.0034 & 0.0025 & 0.0021 & 0.0017 & 0.0015 & 0.0013 & 0.0011 \\
$\delta$, \(R \sim \{2, 3, 4\}\) & 0.0168 & 0.0068 & 0.0051 & 0.0035 & 0.0024 & 0.0017 & 0.0012 \\
$\mathrm{rel}$, fixed R=2 & 0.229 & 0.172 & 0.150 & 0.135 & 0.124 & --- & --- \\
$\mathrm{rel}$, \(R \sim \{1, \dots, 6\}\) & 0.104 & 0.090 & 0.082 & 0.076 & 0.071 & 0.066 & 0.062 \\
$\mathrm{rel}$, \(R \sim \{2, 3, 4\}\) & 0.195 & 0.136 & 0.124 & 0.109 & 0.095 & 0.084 & 0.076 \\
KL, fixed R=2 & 0.085 & 0.038 & 0.028 & 0.021 & 0.017 & --- & --- \\
KL, \(R \sim \{1, \dots, 6\}\) & 0.016 & 0.013 & 0.010 & 0.0075 & 0.0059 & 0.0047 & 0.0040 \\
KL, \(R \sim \{2, 3, 4\}\) & 0.070 & 0.045 & 0.027 & 0.018 & 0.013 & 0.010 & 0.008 \\
\bottomrule
\end{tabular}
\end{table*}

The output-space rows of Table~\ref{tab:3} make the same point without the geometry. Under fixed R=2 the next-token distribution still moves by 0.02--0.03 nats per pass at passes 4--6, while the direction change is down to 0.004--0.006, and it decays slowly (by about $0.75\times$ per pass after pass 3) rather than geometrically: the divergence is not a blow-up but a steady drift of the prediction by a shrinking amount in a direction the loss does not reward. Under \(R \sim \{2, 3, 4\}\) the KL at pass 2 is as large as under fixed R (0.070 against 0.085), so the extra passes change the prediction about as much as a full layer would, and it then contracts faster, to 0.008 by pass 8; under \(R \sim \{1, \dots, 6\}\) it is four to five times smaller from the start, as damped passes should be.

\begin{figure*}[t]
\centering
\includegraphics[width=0.9\textwidth]{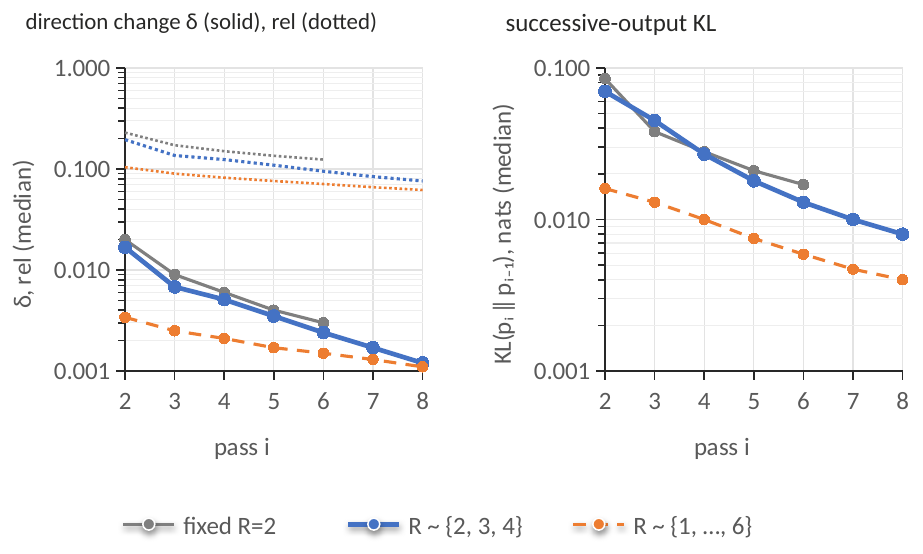}
\caption{Per-pass convergence of the loop, from Table~\ref{tab:3}: direction change $\delta_i$ (left, solid) and relative step $\mathrm{rel}_i$ (left, dotted), both from the seed-42 runs, and successive-output KL (right, seed-43 runs), block-mean of per-block medians on the FineWeb-Edu validation batch at step 60k, log scale. Near-straight lines mean geometric contraction; the rel curves flatten because the norm keeps growing while the direction settles. Fixed R=2 is traced to 6 passes, the sampled-R models to 8.}
\label{fig:conv}
\end{figure*}

Figure~\ref{fig:conv} plots the $\delta$, rel and KL rows of Table~\ref{tab:3} on a logarithmic axis. $\delta$ and KL fall along near-straight lines, i.e. by a roughly constant factor per pass ($\delta$: 0.64 for \(R \sim \{2, 3, 4\}\), 0.83 for \(R \sim \{1, \dots, 6\}\); KL: 0.70 and 0.79), so the per-token iteration is a contraction and the loop approaches a self-consistent state, one that a further pass maps back onto itself up to a tolerance: by pass 8 the direction moves by a cosine of 0.999 and the next-token distribution by under 0.01 nats for both sampled-R models. The level of a curve is the size of the step a pass makes and its slope is the contraction rate, and the two schedules separate on exactly these: \(R \sim \{1, \dots, 6\}\) takes steps five times smaller but contracts more slowly, \(R \sim \{2, 3, 4\}\) starts from larger steps and damps them faster, and the two reach a similar step size by pass 8, each around its own state. The convergence itself is not new: Geiping et al.~\citep{geiping2025scaling}, Logan~\citep{logan2026fixedpoint}, Pappone et al.~\citep{pappone2025twoscale} and Prairie et al.~\citep{prairie2026benice} all report settling across passes. For us it continues an earlier line of work: in 2024 we iterated a transducer joiner to self-consistency against a fixed set of hint embeddings \citep{kolokolov2024selfconsistent}, a loop whose only feedback path is the attention query, and found it converged within three iterations; the looped block studied here is the same self-consistent iteration applied to a full language-model block, and the traces extend that finding to it, with a slower contraction and the convergence in direction rather than norm, which is why the fixed-R loop converges by the same measure while its loss diverges (above), and with the gates setting both the step size and the rate. The dotted rel curves make the norm effect visible in the same panel: under fixed R=2 the relative step flattens at 0.12--0.20 while $\delta$ falls by another order of magnitude, and under the sampled schedules it declines by only about $0.9\times$ per pass, so a norm-based reading would call every loop slow to converge.

Two consequences follow. Norm-based stopping rules are unsafe for a fixed-depth looped model, and sampled-depth training is what makes hidden-state settling a usable proxy for output stability; the step-size exit rule of Pappone et al.~\citep{pappone2025twoscale} is defined on a sampled-depth model and is consistent with this. And the relative step, because it carries the norm, is the wrong metric for diagnosing why a static-depth model fails to share its cache; the direction metric isolates the drift that matters.

\subsection{Forward static KV}\label{sec:statickv}

Reusing the first pass's keys and values on later passes costs 0.012--0.026 nats at the best inference R in every regime (Table~\ref{tab:4}), less at 6 blocks than at 3, constant through training and uniform across domains (Table~\ref{tab:A4}). Training throughput rises by 2\% at fixed R=2, 7\% at mean R=3 and 8\% at mean R=3.5, because only the KV projections of the loop passes are skipped and the 32k-vocabulary head dominates per-token FLOPs at this size. At inference the model decodes with the cache of a plain model of the same block count, which is the same memory saving that backward cache sharing obtains after training (\S\ref{sec:discussion}).

\begin{table*}[t]
\centering
\small
\caption{Validation loss at each model's best inference loop count, and training throughput in tokens per second, for models that recompute keys and values on every pass (fresh) versus models that compute them once on the first pass and reuse them on later passes (static). Both columns are the seed-42 runs, so that each comparison is between two matched single runs; the two-seed means of the fresh 6-block models (Table~\ref{tab:2}) differ from the values here by at most 0.002.}
\label{tab:4}
\vspace{3pt}
\begin{tabular}{lcccc}
\toprule
regime & fresh & static & cost & tok/s \\
\midrule
3 blocks, fixed R=2 & 3.055 & 3.071 & +0.016 & 86.1k $\rightarrow$ 87.6k \\
3 blocks, \(R \sim \{1, \dots, 6\}\) & 3.041 & 3.067 & +0.026 & 57.5k $\rightarrow$ 59.3k \\
6 blocks, fixed R=2 & 2.917 & 2.929 & +0.012 & 51.6k $\rightarrow$ 52.8k \\
6 blocks, \(R \sim \{1, \dots, 6\}\) & 2.916 & 2.933 & +0.017 & 28.5k $\rightarrow$ 30.7k \\
6 blocks, \(R \sim \{2, 3, 4\}\) & 2.908 & 2.924 & +0.016 & 32.4k $\rightarrow$ 34.7k \\
\bottomrule
\end{tabular}
\end{table*}

The dynamics change more than the loss. With static KV the attention output on a later pass is a convex combination of fixed vectors, so a drifting query can only re-weight a fixed set; under fixed R=2 this halves the damage of over-iteration (+0.60 at R=6 vs +1.15), and under \(R \sim \{1, \dots, 6\}\) the per-token iteration is uniformly contractive (ratios 0.7--0.9 in every block, no expansive block) with 44\% of (token, block) pairs with $\delta_2 < 10^{-3}$ against 11\% for fresh KV. The gates also shift: with a fixed memory the attention gate of early blocks drops (block 0 of the 3-block fixed-R model: $0.99 \rightarrow 0.53$) and the second pass's work moves into the FFN. Once R is sampled, the over-iteration robustness advantage mostly disappears, because sampling removes the fragility static KV was protecting against. The price is the lost ability of tokens to read each other's refined states; we did not test tasks that need it.

\subsection{Context: the depth ladder}\label{sec:ladder}

For reference, looping recovers a fraction of what real depth provides. Looping a 3-block model twice improves on the plain 3-layer model by 0.057 nats, 35\% of the 0.161 gap to the plain 6-layer model, at the same FLOPs and throughput (86k vs 89k tok/s). Looping the 6-block model recovers 24\% of the gap to the plain 12-block model (0.032 of 0.133). At equal compute the plain model wins by about 0.10 nats in both pairs, and the fraction recovered falls with depth. The gain is the same size on fineweb, code and math (Table~\ref{tab:A4}).

\begin{table*}[t]
\centering
\small
\caption{Validation loss at the last checkpoint (step 60k, 1.97B training tokens) for the depth ladder. Passes is the number of times each block is applied at inference; looped models are shown at the loop count that gives their lowest validation loss. The plain\_6 row and the two looped\_6 rows are means over seeds 42 and 43 (the seeds differ by at most 0.003); the other rows are single runs.}
\label{tab:5}
\vspace{3pt}
\begin{tabular}{lcccc}
\toprule
model & non-emb params & passes & val loss & tok/s \\
\midrule
plain\_3 & 21.2M & 1 & 3.112 & 131.5k \\
looped\_3, fixed R=2 & 21.2M & 2 & 3.055 & 86.1k \\
looped\_3, \(R \sim \{1, \dots, 6\}\), R=4 & 21.2M & 4 & 3.041 & 57.5k \\
plain\_6 & 42.5M & 1 & 2.951 & 88.9k \\
looped\_6, fixed R=2 & 42.5M & 2 & 2.919 & 51.6k \\
looped\_6, \(R \sim \{2, 3, 4\}\), R=3 & 42.5M & 3 & 2.909 & 32.4k \\
plain\_12 & 85.0M & 1 & 2.818 & 53.6k \\
\bottomrule
\end{tabular}
\end{table*}

\section{Discussion}\label{sec:discussion}

\textbf{Two uses of one loop, and a schedule that keeps both.} A block applied R times can be a deeper network with shared weights or an iteration toward a fixed point, and the training objective decides which. Fixed R produces the former: the second pass is a full layer, with gates that approach 1 in the later blocks, and since training only ever runs and scores the two-pass output, no other depth is ever trained for, so the model is usable only at R=2. Sampling down to R=1 produces the latter: every pass must be safe to skip, so every pass is small, with gates of 0.1--0.5. The \(\{2, 3, 4\}\) schedule shows the two are separable. What the lower bound removes is the demand for a one-pass answer; what the sampling over 2--4 keeps is robustness in that range. For practitioners the recommendation is short: sample R from a range that starts at 2, never at 1, unless a one-pass model is itself a requirement.

\textbf{Relation to recent recipes.} Prairie et al.~\citep{prairie2026benice} obtain convergence through stochastic depth drawn from a Poisson distribution, which includes small depths, and observe that it is necessary but not sufficient for lossless cache sharing at 770M and 1.3B parameters. The gate collapse we see under a lower bound of 1 is a candidate explanation for the insufficiency: a loop whose passes are near-skips has little to share. Jeddi et al.~\citep{jeddi2026loopformer} train on trajectories of varying length with a shortcut-consistency loss that explicitly asks short runs to be usable and long runs to refine; the \(\{2, 3, 4\}\) schedule is the minimal version of that idea, one flag rather than a new objective, and it would be natural to measure where the two differ. Huang et al.~\citep{huang2026optimizers} learn a monotone step size across loops; our constant per-block gates are the simplest member of that family, and their collapse under \(R \sim \{1, \dots, 6\}\) is the model choosing the step size that Huang et al.~would schedule.

\textbf{Direction versus norm.} The pre-norm architecture reads only the direction of the residual stream, and a trained block applied past its trained depth rotates that direction by a shrinking amount each pass while adding to its norm. This defeats both obvious stopping rules, i.e. rules that watch the change between consecutive passes and stop looping once it is small. A norm-based rule (the relative step $\mathrm{rel}_i$, or any measure that includes the vector's length) never fires under the model trained with fixed R=2, because the norm grows about 9\% per pass and the step floors at 0.12--0.20; it reports non-convergence while the direction has in fact settled. A direction-based rule ($\delta_i$) fires by pass 4--5, and is right about the hidden state, but the loss over those passes rises from 2.92 to 4.06, because the direction is settling toward a state the output projection never saw; it reports convergence while the predictions are getting worse. Neither signal, taken from the hidden state alone, says whether another pass helps. A stopping rule for a looped model therefore has to be stated in output space (the successive-output KL of Table~\ref{tab:3}, or the loss itself), or the model has to be trained so that hidden-state settling and output stability coincide, which sampled depth does: under the model trained with \(R \sim \{2, 3, 4\}\) the loss is flat over the range in which the direction settles.

\textbf{Forward versus backward KV.} Freezing the first pass's memory decouples the per-token fixed points and makes the iteration contractive, at a small and regime-independent cost; broadcasting the last pass's memory backward after training recovers the same cache saving, zero-shot when pretraining sampled the depth and with a short finetuning stage otherwise. The forward variant is a training-time choice with a dynamics benefit, the backward one a post-training choice that leaves pretraining unchanged; which is preferable depends on whether the model is being trained anyway.

\textbf{What the loop buys.} At equal parameters, about a third of what doubling the depth provides; at equal FLOPs, nothing, on language-modeling loss at this scale. The case for looping rests on parameter efficiency and on gains that perplexity does not measure, which is consistent with how \citep{geiping2025scaling,zhu2025ouro,marchenko2026closing,huang2026optimizers} report their results.

\textbf{Future work: what to loop.} Every looped language model we know of, including ours, iterates the full block; where adapters appear they ride on the looped block \citep{bae2024relaxed}, install the loop in a pretrained one \citep{shapiro2026retrofitting}, or select which tokens receive extra full-backbone passes \citep{fu2026thinkathard}. Our gates point at a cheaper loop body: in the early and middle blocks of every fresh-KV looped model the attention gate exceeds the FFN gate (Table~\ref{tab:A5}; the block means are 0.71 vs 0.63 under fixed R=2 and 0.65 vs 0.62 under \(R \sim \{2, 3, 4\}\), with the last block the exception at 0.5--0.7 vs 0.9), so re-reading the sequence with a refined query was worth more than re-running the per-token FFN, which is also the largest single cost of a pass; under static KV the ordering reverses, as \S\ref{sec:statickv} showed. We therefore intend to compare three loop bodies, i.e. three choices of what runs on passes 2 to R while pass 1 stays a full block, each with fresh and with static KV, under the \(\{2, 3, 4\}\) schedule, on loss against inference FLOPs. The first is the full block studied here: attention and FFN on every pass. The second keeps attention but computes the FFN only on the first pass; on later passes a small low-rank adapter takes the FFN's place and the first pass's FFN output is added back unchanged, which makes a loop pass cost roughly 40\% of a full one. The third drops attention as well and runs only the adapter, so a pass is a per-token refinement with no cross-token communication, at about 1\% of a full pass's cost. Dong et al.~\citep{dong2021attention} showed that pure attention collapses in rank and that the FFN is what prevents it, so the adapter is the minimal nonlinearity that should keep the attention-only loop from over-smoothing, and the re-injected first-pass output is the anchor that made the self-consistent joiner of \citep{kolokolov2024selfconsistent} converge in three iterations. The \(R \sim \{2, \dots, 6\}\) control for \S\ref{sec:gates}, deep supervision at intermediate passes, looping only the deeper blocks, and a replication at 150M--600M parameters with downstream tasks complete the list.

\section{Limitations}\label{sec:limitations}

All models have 21--85M non-embedding parameters and were trained on 2B tokens on one consumer GPU; the effects are 0.01--0.1 nats and we make no claim about scale. The headline comparison (the plain 6-block model and the looped 6-block model under each of the three schedules) has two seeds, which agree to within 0.003 nats at every trained loop count and give the same gate profiles; every other run is single-seed, and the pattern's consistency across two depths, three domains and both KV policies is our evidence that it is not noise. We evaluate language-modeling loss only, so we cannot say whether the shallow test-time scaling reflects the metric or the models. The block has no input injection, step embedding or truncated backpropagation, which keeps the comparison minimal but differs from the recipes of \citep{geiping2025scaling,zhu2025ouro,marchenko2026closing}; all blocks loop with the same R although the gates suggest the first block should not. The convergence analysis is in hidden-state space; the output-space metric (the successive-output KL of Table~\ref{tab:3}) is reported for one seed. We did not test tasks that require tokens to read each other's refined states, which is where forward static KV should cost most.

\section{Conclusion}\label{sec:conclusion}

What matters in a loop-count schedule is its smallest value of iteration count, not the fact that R is sampled. If training ever runs the block only once, every extra pass learns to be a small correction (the gates settle at 0.1--0.5); if every training step runs the block at least twice, each extra pass learns to act as a full layer (gates of 0.6--0.9). The learned gates make this visible, and changing just the lower bound of the sampling range from 1 to 2 is enough to switch between the two regimes. A model trained with R sampled from \(\{2, 3, 4\}\) matches the quality of the model trained at fixed R=2 and, unlike it, can be run at any R in its trained range and beyond without loss.

The loop converges, and it converges in direction: with every pass the residual stream rotates by a shrinking amount while its norm keeps growing, and since a pre-norm block reads only the direction, the iteration approaches a self-consistent state, one that a further pass maps back onto itself, in the same way that the self-consistent joiner of our earlier work \citep{kolokolov2024selfconsistent} settled against a fixed memory. With sampled R the state settles and the loss stays flat; with fixed R the state settles toward a point the output layer never saw, so the loss gets worse. A fixed-R model therefore cannot tell from its hidden state when to stop looping.

Forward static KV, which freezes the first pass's keys and values for the later passes, costs 0.01--0.03 nats in every regime, makes the iteration uniformly contractive, halves the inference cache, and is complementary to backward sharing of the final pass's cache after training. These are measured claims at small scale, released with their runs and traces so they can be checked at larger ones.

\bibliographystyle{unsrtnat}
\bibliography{refs}

\clearpage
\onecolumn
\appendix
\setcounter{table}{0}\renewcommand{\thetable}{A\arabic{table}}
\renewcommand{\thesection}{Appendix \Alph{section}:}
\section{Full run tables}\label{sec:appendix}

\begin{table*}[htbp]
\centering
\footnotesize\setlength{\tabcolsep}{3pt}
\caption{All thirteen seed-42 runs: validation loss at each inference loop count R, the best value, and training throughput in tokens per second.}
\label{tab:A1}
\vspace{3pt}
\begin{tabular}{lccccccccc}
\toprule
run & R=1 & R=2 & R=3 & R=4 & R=5 & R=6 & R=8 & best & tok/s \\
\midrule
plain\_3 & 3.112 &  &  &  &  &  &  & 3.112 & 131.5k \\
looped\_3, fixed R=2 & 5.685 & 3.055 & 3.288 & 3.668 &  & 4.970 &  & 3.055 & 86.1k \\
looped\_3\_skv, fixed R=2 & 4.737 & 3.071 & 3.259 & 3.512 &  & 4.288 &  & 3.071 & 87.6k \\
looped\_3\_rs, \(R \sim \{1, \dots, 6\}\) & 3.229 & 3.079 & 3.048 & 3.041 & 3.041 & 3.045 & 3.059 & 3.041 & 57.5k \\
looped\_3\_skv\_rs, \(R \sim \{1, \dots, 6\}\) & 3.196 & 3.092 & 3.072 & 3.067 & 3.068 & 3.072 & 3.084 & 3.067 & 59.3k \\
plain\_6 & 2.951 &  &  &  &  &  &  & 2.951 & 88.9k \\
looped\_6, fixed R=2 & 5.046 & \textbf{2.917} & 3.007 & 3.142 & --- & 4.063 & --- & 2.917 & 51.6k \\
looped\_6\_skv, fixed R=2 & 5.012 & 2.929 & 3.026 & 3.162 &  & 3.527 &  & 2.929 & 52.8k \\
looped\_6\_rs, \(R \sim \{1, \dots, 6\}\) & 3.002 & 2.933 & 2.919 & 2.916 & 2.918 & 2.922 & 2.932 & 2.916 & 28.5k \\
looped\_6\_skv\_rs, \(R \sim \{1, \dots, 6\}\) & 2.989 & 2.946 & 2.935 & 2.933 & 2.935 & 2.938 & 2.947 & 2.933 & 30.7k \\
looped\_6\_r234, \(R \sim \{2, 3, 4\}\) & 4.557 & 2.927 & 2.908 & 2.910 & 2.917 & 2.926 & 2.945 & 2.908 & 32.4k \\
looped\_6\_skv\_r234, \(R \sim \{2, 3, 4\}\) & 3.945 & 2.938 & 2.924 & 2.926 & 2.933 & 2.942 & 2.961 & 2.924 & 34.7k \\
plain\_12 & 2.818 &  &  &  &  &  &  & 2.818 & 53.6k \\
\bottomrule
\end{tabular}
\end{table*}

\begin{table*}[htbp]
\centering
\footnotesize\setlength{\tabcolsep}{3pt}
\caption{Seed-43 repeats of the four headline runs, same columns as Table A1.}
\label{tab:A2}
\vspace{3pt}
\begin{tabular}{lccccccccc}
\toprule
run & R=1 & R=2 & R=3 & R=4 & R=5 & R=6 & R=8 & best & tok/s \\
\midrule
plain\_6\_s43 & 2.952 &  &  &  &  &  &  & 2.952 & 89.0k \\
looped\_6\_s43, fixed R=2 & 4.866 & \textbf{2.920} & 3.016 & 3.138 & --- & 3.434 & --- & 2.920 & 50.9k \\
looped\_6\_rs\_s43, \(R \sim \{1, \dots, 6\}\) & 3.001 & 2.932 & 2.918 & \textbf{2.915} & 2.917 & 2.921 & 2.932 & 2.915 & 30.7k \\
looped\_6\_r234\_s43, \(R \sim \{2, 3, 4\}\) & 4.647 & 2.928 & \textbf{2.910} & 2.912 & 2.918 & 2.927 & 2.948 & 2.910 & 34.7k \\
\bottomrule
\end{tabular}
\end{table*}

\begin{table*}[htbp]
\centering
\footnotesize\setlength{\tabcolsep}{3pt}
\caption{Growth of the residual norm per loop pass: median over tokens of $\lVert s_i \rVert / \lVert s_{i-1} \rVert$ within each block, averaged over the six blocks, for the seed-42 6-block fresh-KV models on the same tokens as Table~\ref{tab:3}. Pass 1, the ungated first pass, is omitted, as in as Table~\ref{tab:3}. Over passes 2 to R the per-block values lie within 1.02--1.24 under fixed R=2 and 1.02--1.10 under the sampled schedules.}
\label{tab:A3}
\vspace{3pt}
\begin{tabular}{lccccccc}
\toprule
schedule & p2 & p3 & p4 & p5 & p6 & p7 & p8 \\
\midrule
fixed R=2 & 1.094 & 1.095 & 1.093 & 1.092 & 1.089 & --- & --- \\
\(R \sim \{1, \dots, 6\}\) & 1.058 & 1.052 & 1.049 & 1.047 & 1.045 & 1.044 & 1.042 \\
\(R \sim \{2, 3, 4\}\) & 1.089 & 1.074 & 1.076 & 1.073 & 1.067 & 1.061 & 1.057 \\
\bottomrule
\end{tabular}
\end{table*}

\begin{table*}[htbp]
\centering
\footnotesize\setlength{\tabcolsep}{3pt}
\caption{Per-domain validation loss at each model's best inference loop count R for every run (seed 42; the four headline runs also seed 43, shown as 42 / 43). The static-KV cost is 0.01--0.03 nats in every domain, and the looping gain is the same fraction of the plain-depth gap in each domain.}
\label{tab:A4}
\vspace{3pt}
\begin{tabular}{lcccc}
\toprule
model & best R & fineweb & code & math \\
\midrule
plain\_3 & 1 & 3.321 & 1.868 & 2.688 \\
plain\_6 & 1 & 3.159 / 3.160 & 1.722 / 1.719 & 2.516 / 2.517 \\
plain\_12 & 1 & 3.025 & 1.605 & 2.379 \\
looped\_3, fixed R=2 & 2 & 3.264 & 1.815 & 2.625 \\
looped\_3\_skv, fixed R=2 & 2 & 3.279 & 1.836 & 2.647 \\
looped\_3\_rs, \(R \sim \{1, \dots, 6\}\) & 4 & 3.248 & 1.807 & 2.620 \\
looped\_3\_skv\_rs, \(R \sim \{1, \dots, 6\}\) & 4 & 3.274 & 1.837 & 2.646 \\
looped\_6, fixed R=2 & 2 & 3.126 / 3.128 & 1.691 / 1.692 & 2.479 / 2.484 \\
looped\_6\_skv, fixed R=2 & 2 & 3.136 & 1.707 & 2.495 \\
looped\_6\_r234, \(R \sim \{2, 3, 4\}\) & 3 & 3.116 / 3.117 & 1.684 / 1.684 & 2.472 / 2.472 \\
looped\_6\_skv\_r234, \(R \sim \{2, 3, 4\}\) & 3 & 3.131 & 1.701 & 2.487 \\
looped\_6\_rs, \(R \sim \{1, \dots, 6\}\) & 4 & 3.124 / 3.122 & 1.694 / 1.694 & 2.481 / 2.484 \\
looped\_6\_skv\_rs, \(R \sim \{1, \dots, 6\}\) & 4 & 3.141 & 1.707 & 2.497 \\
\bottomrule
\end{tabular}
\end{table*}

\begin{table*}[htbp]
\centering
\scriptsize\setlength{\tabcolsep}{3pt}
\caption{Learned gates $\alpha_\mathrm{attn}$ / $\alpha_\mathrm{ff}$ per block at step 60k for every looped run (Table~\ref{tab:1} in the body is the two-seed mean of the 6-block fresh-KV rows).}
\label{tab:A5}
\vspace{3pt}
\begin{tabular}{lcccccc}
\toprule
run & block 0 & block 1 & block 2 & block 3 & block 4 & block 5 \\
\midrule
looped\_3, fixed R=2 & 0.99 / 0.29 & 0.98 / 0.70 & 1.00 / 0.91 &  &  &  \\
looped\_3\_skv, fixed R=2 & 0.53 / 0.38 & 0.79 / 0.79 & 1.00 / 0.94 &  &  &  \\
looped\_3\_rs, \(R \sim \{1, \dots, 6\}\) & 0.59 / 0.17 & 0.58 / 0.54 & 0.58 / 0.72 &  &  &  \\
looped\_3\_skv\_rs, \(R \sim \{1, \dots, 6\}\) & 0.47 / 0.16 & 0.32 / 0.45 & 0.81 / 0.68 &  &  &  \\
looped\_6, fixed R=2, seed 42 & 0.28 / 0.15 & 0.50 / 0.40 & 0.81 / 0.65 & 0.98 / 0.72 & 1.00 / 0.96 & 0.67 / 0.91 \\
looped\_6, fixed R=2, seed 43 & 0.34 / 0.22 & 0.58 / 0.45 & 0.89 / 0.55 & 0.96 / 0.69 & 1.00 / 0.96 & 0.59 / 0.91 \\
looped\_6\_rs, \(R \sim \{1, \dots, 6\}\), seed 42 & 0.38 / 0.09 & 0.31 / 0.27 & 0.33 / 0.40 & 0.40 / 0.44 & 0.44 / 0.39 & 0.12 / 0.40 \\
looped\_6\_rs, \(R \sim \{1, \dots, 6\}\), seed 43 & 0.44 / 0.11 & 0.30 / 0.33 & 0.33 / 0.35 & 0.40 / 0.45 & 0.47 / 0.35 & 0.11 / 0.35 \\
looped\_6\_r234, \(R \sim \{2, 3, 4\}\), seed 42 & 0.43 / 0.16 & 0.47 / 0.47 & 0.66 / 0.56 & 0.81 / 0.77 & 1.00 / 0.83 & 0.52 / 0.90 \\
looped\_6\_r234, \(R \sim \{2, 3, 4\}\), seed 43 & 0.51 / 0.18 & 0.48 / 0.56 & 0.65 / 0.55 & 0.80 / 0.73 & 1.00 / 0.82 & 0.46 / 0.88 \\
looped\_6\_skv, fixed R=2 & 0.10 / 0.19 & 0.41 / 0.42 & 0.47 / 0.83 & 0.68 / 0.81 & 1.00 / 0.89 & 0.93 / 0.95 \\
looped\_6\_skv\_rs, \(R \sim \{1, \dots, 6\}\) & 0.12 / 0.08 & 0.10 / 0.15 & 0.12 / 0.26 & 0.20 / 0.31 & 0.54 / 0.32 & 0.18 / 0.40 \\
looped\_6\_skv\_r234, \(R \sim \{2, 3, 4\}\) & 0.21 / 0.16 & 0.27 / 0.43 & 0.32 / 0.66 & 0.47 / 0.66 & 1.00 / 0.75 & 0.69 / 0.93 \\
\bottomrule
\end{tabular}
\end{table*}

\begin{table*}[htbp]
\centering
\footnotesize\setlength{\tabcolsep}{3pt}
\caption{Fraction of (token, block) pairs whose direction change $\delta_i$ is below $10^{-3}$ at each pass, FineWeb-Edu validation batch, step 60k.}
\label{tab:A6}
\vspace{3pt}
\begin{tabular}{lccccccc}
\toprule
run & p2 & p3 & p4 & p5 & p6 & p7 & p8 \\
\midrule
looped\_6\_rs, seed 42 & 0.11 & 0.25 & 0.34 & 0.43 & 0.50 & 0.57 & 0.65 \\
looped\_6\_rs, seed 43 & 0.10 & 0.27 & 0.36 & 0.44 & 0.50 & 0.56 & 0.62 \\
looped\_6\_skv\_rs & 0.44 & 0.48 & 0.54 & 0.59 & 0.63 & 0.71 & 0.80 \\
looped\_6\_r234, seed 42 & 0.00 & 0.03 & 0.10 & 0.24 & 0.36 & 0.42 & 0.50 \\
looped\_6\_r234, seed 43 & 0.01 & 0.07 & 0.15 & 0.26 & 0.37 & 0.46 & 0.55 \\
looped\_6\_skv\_r234 & 0.05 & 0.13 & 0.21 & 0.29 & 0.38 & 0.46 & 0.52 \\
\bottomrule
\end{tabular}
\end{table*}

\end{document}